\documentclass[runningheads,oribibl]{llncs}
\usepackage{caption}
\usepackage[switch]{lineno}
\usepackage[ruled,vlined]{algorithm2e}
\usepackage{mathtools}
\usepackage[flushleft]{threeparttable}
\usepackage{booktabs}
\usepackage{enumerate}
\usepackage{enumitem}
\usepackage{xcolor}
\usepackage{xspace}
\usepackage{graphicx}
\usepackage{microtype}
\usepackage{multirow}
\usepackage{wrapfig}
\usepackage[misc,geometry]{ifsym}

\usepackage{listings}
\definecolor{codegreen}{rgb}{0,0.6,0}
\definecolor{codegray}{rgb}{0.5,0.5,0.5}
\definecolor{codepurple}{rgb}{0.58,0,0.82}
\definecolor{backcolour}{rgb}{0.95,0.95,0.92}
\lstdefinestyle{mystyle}{
    backgroundcolor=\color{backcolour},   
    commentstyle=\color{codegreen},
    keywordstyle=\color{magenta},
    numberstyle=\tiny\color{codegray},
    stringstyle=\color{codepurple},
    basicstyle=\ttfamily\footnotesize,
    breakatwhitespace=false,         
    breaklines=false,                 
    captionpos=b,                    
    keepspaces=true,                 
    numbers=left,                    
    numbersep=5pt,                  
    showspaces=false,                
    showstringspaces=false,
    showtabs=false,                  
    tabsize=2
}
\definecolor{mint}{rgb}{0.24, 0.71, 0.54}
\newcommand{\lp}{\texttt{LayoutParser}\xspace}

\begin{document}
\title{\lp: A Unified Toolkit for Deep Learning Based Document Image Analysis} 
\titlerunning{\lp: A Unified Toolkit for DL-Based DIA}

%
%
\author{Zejiang Shen\inst{1} (\Letter)\and
Ruochen Zhang\inst{2} \and
Melissa Dell\inst{3} \and 
Benjamin Charles Germain Lee\index{Lee, Benjamin}\inst{4} \and 
Jacob Carlson\inst{3} \and
Weining Li\inst{5}}
\authorrunning{Z. Shen et al.}
%
\institute{Allen Institute for AI\\
\email{shannons@allenai.org}\\
\and
Brown University\\
\email{ruochen\_zhang@brown.edu}\\
\and
Harvard University\\
\email{\{melissadell,jacob\_carlson\}@fas.harvard.edu}\\
\and 
University of Washington\\
\email{bcgl@cs.washington.edu}\\
\and
University of Waterloo\\
\email{w422li@uwaterloo.ca}
}

\maketitle              
\begin{abstract}

Recent advances in document image analysis (DIA) have been primarily driven by the application of neural networks. 
Ideally, research outcomes could be easily deployed in production and extended for further investigation. 
However, various factors like loosely organized codebases and sophisticated model configurations complicate the easy reuse of important innovations by a wide audience. 
Though there have been on-going efforts to improve reusability and simplify deep learning (DL) model development in disciplines like natural language processing and computer vision, none of them are optimized for challenges in the domain of DIA. This represents a major gap in the existing toolkit, as DIA is central to academic research across a wide range of disciplines in the social sciences and humanities. 
This paper introduces \lp, an open-source library for streamlining the usage of DL in DIA research and applications. 
The core \lp library comes with a set of simple and intuitive interfaces for applying and customizing DL models for layout detection, character recognition, and many other document processing tasks. 
To promote extensibility, \lp also incorporates a community platform for sharing both pre-trained models and full document digitization pipelines. We demonstrate that \lp is helpful for both lightweight and large-scale digitization pipelines in real-word use cases. 
The library is publicly available at \url{https://layout-parser.github.io}. 

\end{abstract}

\section{Introduction}

Deep Learning(DL)-based approaches are the state-of-the-art for a wide range of document image analysis (DIA) tasks including document image classification~\cite{harley2015evaluation, xu2019layoutlm}, layout detection~\cite{zhong2019publaynet, oliveira2018dhsegment}, table detection~\cite{prasad2020cascadetabnet}, and scene text detection~\cite{baek2019character}.
A generalized learning-based framework dramatically reduces the need for the manual specification of complicated rules, which is the status quo with traditional methods. 
DL has the potential to transform DIA pipelines and benefit a broad spectrum of large-scale document digitization projects. 

However, there are several practical difficulties for taking advantages of recent advances in DL-based methods:
1) DL models are notoriously convoluted for reuse and extension. 
Existing models are developed using distinct frameworks like TensorFlow~\cite{tensorflow2015-whitepaper} or PyTorch~\cite{paszke2019pytorch}, and the high-level parameters can be obfuscated by implementation details~\cite{gardner2018allennlp}. 
It can be a time-consuming and frustrating experience to debug, reproduce, and adapt existing models for DIA, and \textit{many researchers who would benefit the most from using these methods lack the technical background to implement them from scratch.} 
2) Document images contain diverse and disparate patterns across domains, and customized training is often required to achieve a desirable detection accuracy. 
Currently \textit{there is no full-fledged infrastructure for easily curating the target document image datasets and fine-tuning or re-training the models.} 
3) DIA usually requires a sequence of models and other processing to obtain the final outputs. 
Often research teams use DL models and then perform further document analyses in separate processes, and these pipelines are not documented in any central location (and often not documented at all). This makes it \textit{difficult for research teams to learn about how full pipelines are implemented} and \textit{leads them to invest significant resources in reinventing the DIA wheel}. 

\lp provides a unified toolkit to support DL-based document image analysis and processing. 
To address the aforementioned challenges, \lp is built with the following components: 
\begin{enumerate}
    \item An off-the-shelf toolkit for applying DL models for layout detection, character recognition, and other DIA tasks (Section~\ref{sec:core-library})
    \item A rich repository of pre-trained neural network models (Model Zoo) that underlies the off-the-shelf usage
    \item Comprehensive tools for efficient document image data annotation and model tuning to support different levels of customization
    \item A DL model hub and community platform for the easy sharing, distribution, and discussion of DIA models and pipelines, to promote reusability, reproducibility, and extensibility (Section~\ref{sec:community-platform})
\end{enumerate}
The library implements simple and intuitive Python APIs without sacrificing generalizability and versatility, and can be easily installed via pip. 
Its convenient functions for handling document image data can be seamlessly integrated with existing DIA pipelines. 
With detailed documentations and carefully curated tutorials, we hope this tool will benefit a variety of end-users, and will lead to advances in applications in both industry and academic research. 

\lp is well aligned with recent efforts for improving DL model reusability in other disciplines like natural language processing~\cite{gardner2018allennlp, wolf2019huggingface} and computer vision~\cite{wu2019detectron2}, but with a focus on unique challenges in DIA.
We show \lp can be applied in sophisticated and large-scale digitization projects that require precision, efficiency, and robustness, as well as simple and light-weight document processing tasks focusing on efficacy and flexibility (Section~\ref{sec:use-case}). 
\lp is being actively maintained, and support for more deep learning models and novel methods in text-based layout analysis methods~\cite{xu2019layoutlm, wolf2019huggingface} is planned.

The rest of the paper is organized as follows. Section~\ref{sec:related-work} provides an overview of related work.
The core \lp library, DL Model Zoo, and customized model training are described in Section~\ref{sec:core-library}, and the DL model hub and community platform are detailed in Section~\ref{sec:community-platform}. 
Section~\ref{sec:use-case} shows two examples of how \lp can be used in practical DIA projects, and Section~\ref{sec:conclusion} concludes. 

\section{Related Work}
\label{sec:related-work}

Recently, various DL models and datasets have been developed for layout analysis tasks. 
The dhSegment~\cite{oliveira2018dhsegment} utilizes fully convolutional networks~\cite{long2015fully} for segmentation tasks on historical documents. 
Object detection-based methods like Faster R-CNN~\cite{ren2015faster} and Mask R-CNN~\cite{he2017mask} are used for identifying document elements~\cite{zhong2019publaynet} and detecting tables~\cite{schreiber2017deepdesrt,prasad2020cascadetabnet}. 
Most recently, Graph Neural Networks~\cite{scarselli2008graph} have also been used in table detection~\cite{qasim2019rethinking}.
However, these models are usually implemented individually and there is no unified framework to load and use such models. 

There has been a surge of interest in creating open-source tools for document image processing: a search of \verb+document image analysis+ in Github leads to 5M relevant code pieces~\footnote{The number shown is obtained by specifying the search type as `code'.}; yet most of them rely on traditional rule-based methods or provide limited functionalities. 
The closest prior research to our work is the OCR-D project\footnote{https://ocr-d.de/en/about}, which also tries to build a complete toolkit for DIA. 
However, similar to the platform developed by Neudecker et al.~\cite{neudecker2011experimental}, it is designed for analyzing historical documents, and provides no supports for recent DL models.
The \texttt{DocumentLayoutAnalysis} project\footnote{https://github.com/BobLd/DocumentLayoutAnalysis} focuses on processing born-digital PDF documents via analyzing the stored PDF data. 
Repositories like \texttt{DeepLayout}\footnote{https://github.com/leonlulu/DeepLayout} and \texttt{Detectron2-PubLayNet}\footnote{https://github.com/hpanwar08/detectron2} are individual deep learning models trained on layout analysis datasets without support for the full DIA pipeline.
The Document Analysis and Exploitation (DAE) platform~\cite{lamiroy2011open} and the DeepDIVA project~\cite{alberti2018deepdiva} aim to improve the reproducibility of DIA methods (or DL models), yet they are not actively maintained.
OCR engines like \texttt{Tesseract} \cite{tesseract}, \texttt{easyOCR}\footnote{https://github.com/JaidedAI/EasyOCR} and \texttt{paddleOCR}\footnote{https://github.com/PaddlePaddle/PaddleOCR} usually do not come with comprehensive functionalities for other DIA tasks like layout analysis. 

Recent years have also seen numerous efforts to create libraries for promoting reproducibility and reusability in the field of DL. Libraries like Dectectron2~\cite{wu2019detectron2}, AllenNLP~\cite{gardner2018allennlp} and transformers~\cite{wolf2019huggingface} have provided the community with complete DL-based support for developing and deploying models for general computer vision and natural language processing problems. \lp, on the other hand, specializes specifically in DIA tasks. \lp is also equipped with a community platform inspired by established model hubs such as \texttt{Torch Hub}~\cite{paszke2017automatic} and \texttt{TensorFlow Hub}~\cite{tensorflow2015-whitepaper}. It enables the sharing of pretrained models as well as full document processing pipelines that are unique to DIA tasks.

There have been a variety of document data collections to facilitate the development of DL models. Some examples include PRImA~\cite{antonacopoulos2009realistic}(magazine layouts), PubLayNet~\cite{zhong2019publaynet}(academic paper layouts), Table Bank~\cite{li2019tablebank}(tables in academic papers), Newspaper Navigator Dataset~\cite{newspaper_navigator_search_application,newspaper_navigator_dataset}(newspaper figure layouts) and \texttt{HJDataset} \cite{shen2020large}(historical Japanese document layouts). A spectrum of models trained on these datasets are currently available in the \lp model zoo to support different use cases. 

\begin{figure}[t]
    \centering
    \includegraphics[width=0.8\linewidth]{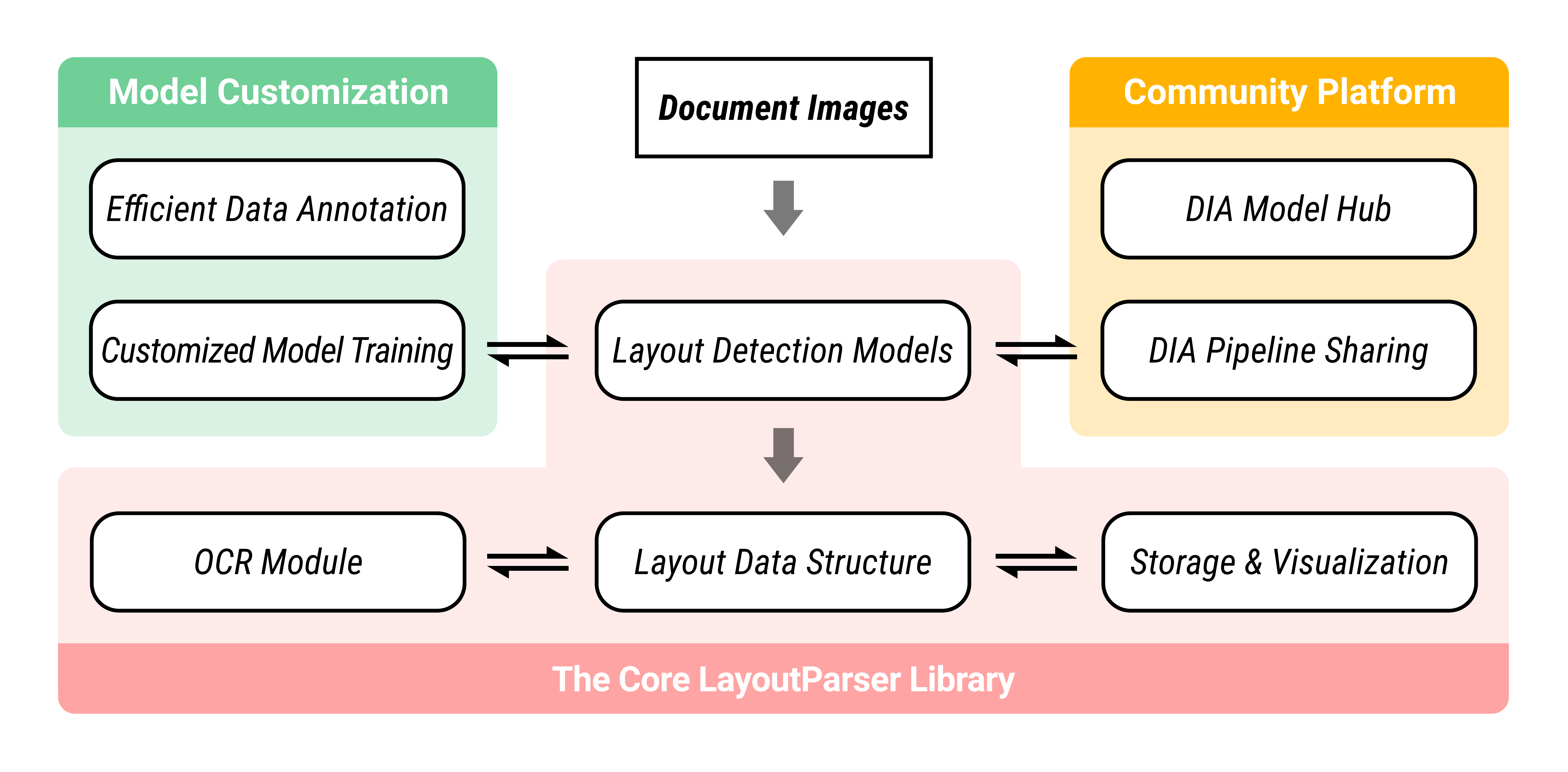}
    \caption{The overall architecture of \lp. For an input document image, the core \lp library provides a set of off-the-shelf tools for layout detection, OCR, visualization, and storage, backed by a carefully designed layout data structure. \lp also supports high level customization via efficient layout annotation and model training functions. These improve model accuracy on the target samples. The community platform enables the easy sharing of DIA models and whole digitization pipelines to promote reusability and reproducibility. A collection of detailed documentation, tutorials and exemplar projects make \lp easy to learn and use.}
    \label{fig:overall}
  \end{figure}
\section{The Core \lp Library}
\label{sec:core-library}

At the core of \lp is an off-the-shelf toolkit that streamlines DL-based document image analysis. 
Five components support a simple interface with comprehensive functionalities:
1) The \emph{layout detection models} enable using pre-trained or self-trained DL models for layout detection with just four lines of code. 
2) The detected layout information is stored in carefully engineered \emph{layout data structures}, which are optimized for efficiency and versatility. 
3) When necessary, users can employ existing or customized OCR models via the unified API provided in the \emph{OCR module}. 
4) \lp comes with a set of utility functions for the \emph{visualization and storage} of the layout data. 
5) \lp is also highly customizable, via its integration with functions for \emph{layout data annotation and model training}.
We now provide detailed descriptions for each component. 

\begin{table}[t]
    
    \centering
    \caption{Current layout detection models in the \lp model zoo}
    \label{table:models}
    \resizebox{\linewidth}{!}{
    \begin{threeparttable}
        \setlength{\tabcolsep}{0.5em} 
        \renewcommand{\arraystretch}{1.2}
        \begin{tabular}{c|c|c|l}
            \toprule
            \textbf{Dataset}                             & \textbf{Base Model}\tnote{1} & \textbf{Large Model} & \textbf{Notes} \\
            \midrule
            PubLayNet~\cite{zhong2019publaynet}          & F / M               & M                    & Layouts of modern scientific documents \\
            PRImA~\cite{antonacopoulos2009realistic}     & M                   & -                    & Layouts of scanned modern magazines and scientific reports \\
            Newspaper~\cite{newspaper_navigator_dataset} & F                   & -                    & Layouts of scanned US newspapers from the 20th century  \\
            TableBank~\cite{li2019tablebank}             & F                   & F                    & Table region on modern scientific and business document    \\
            HJDataset~\cite{shen2020large}               & F / M            & -                    & Layouts of history Japanese documents   \\                  
            \bottomrule      
        \end{tabular}
        \begin{tablenotes}
            \item[1] For each dataset, we train several models of different sizes for different needs (the trade-off between accuracy vs. computational cost). For ``base model'' and ``large model'', we refer to using the ResNet 50 or ResNet 101 backbones~\cite{he2016deep}, respectively. One can train models of different architectures, like Faster R-CNN~\cite{ren2015faster} (F) and Mask R-CNN~\cite{he2017mask} (M). For example, an F in the Large Model column indicates it has a Faster R-CNN model trained using the ResNet 101 backbone. The platform is maintained and a number of additions will be made to the model zoo in coming months.
        \end{tablenotes}
    \end{threeparttable}
    }
\end{table}

\subsection{Layout Detection Models} 

In \lp, a layout model takes a document image as an input and generates a list of rectangular boxes for the target content regions. 
Different from traditional methods, it relies on deep convolutional neural networks rather than manually curated rules to identify content regions.
It is formulated as an object detection problem and state-of-the-art models like Faster R-CNN~\cite{ren2015faster} and Mask R-CNN~\cite{he2017mask} are used. 
This yields prediction results of high accuracy and makes it possible to build a concise, generalized interface for layout detection. 
\lp, built upon Detectron2~\cite{wu2019detectron2}, provides a minimal API that can perform layout detection with only four lines of code in Python:

\begin{lstlisting}[language=python]
import layoutparser as lp
image = cv2.imread("image_file") # load images
model = lp.Detectron2LayoutModel(
    "lp://PubLayNet/faster_rcnn_R_50_FPN_3x/config")
layout = model.detect(image)
\end{lstlisting}
     
\lp provides a wealth of pre-trained model weights using various datasets covering different languages, time periods, and document types. 
Due to domain shift~\cite{ganin2015unsupervised}, the prediction performance can notably drop when models are applied to target samples that are significantly different from the training dataset. 
As document structures and layouts vary greatly in different domains, it is important to select models trained on a dataset similar to the test samples.
A semantic syntax is used for initializing the model weights in \lp, using both the dataset name and model name \verb+lp://<dataset-name>/<model-architecture-name>+.
Shown in Table~\ref{table:models}, \lp currently hosts 9 pre-trained models trained on 5 different datasets. 
Description of the training dataset is provided alongside with the trained models such that users can quickly identify the most suitable models for their tasks.
Additionally, when such a model is not readily available, \lp also supports training customized layout models and community sharing of the models (detailed in Section~\ref{sec:training}).

\begin{figure}[t]
    \centering
    \includegraphics[width=0.65\linewidth]{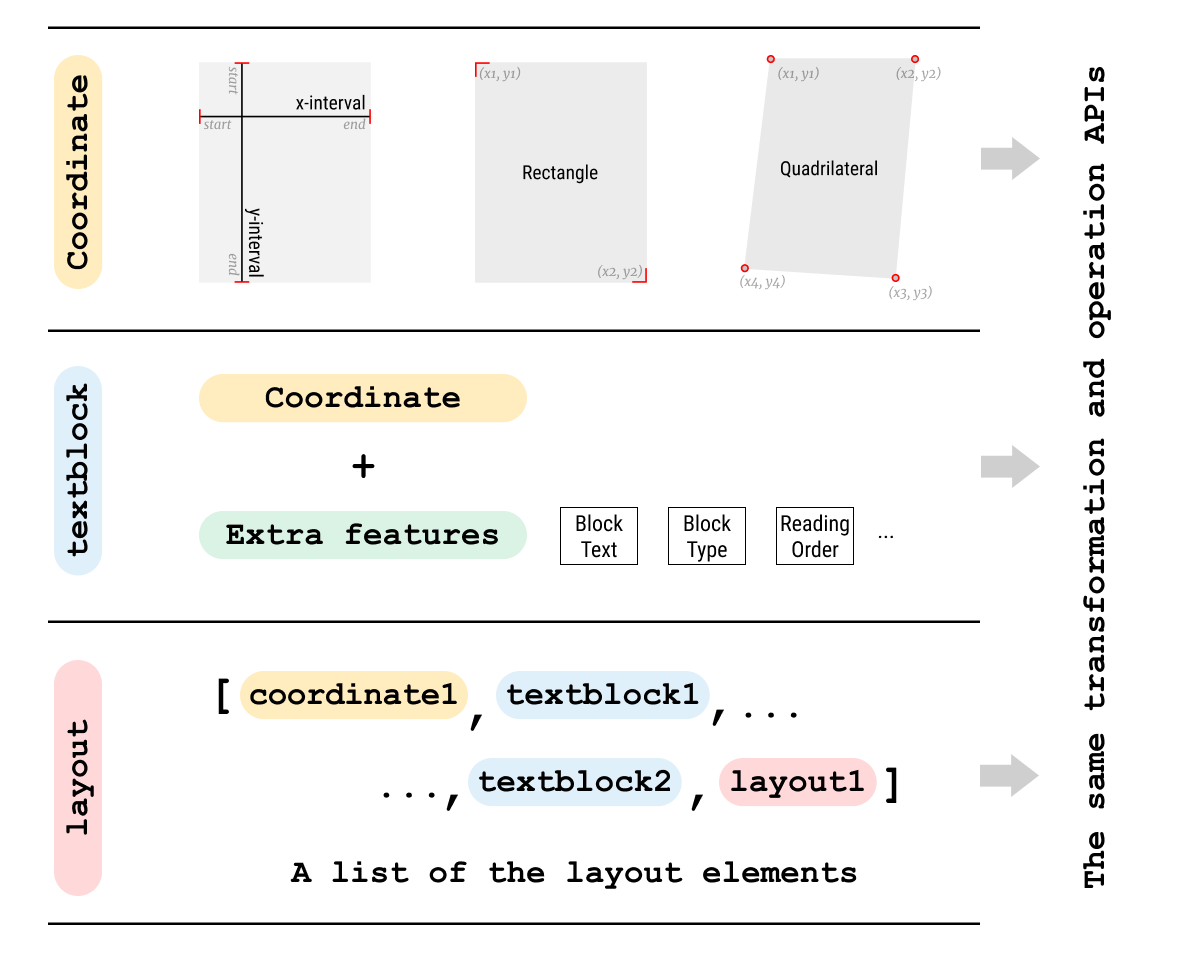}
    \caption{The relationship between the three types of layout data structures. \texttt{Coordinate} supports three kinds of variation; \texttt{TextBlock} consists of the coordinate information and extra features like block text, types, and reading orders; a \texttt{Layout} object is a list of all possible layout elements, including other \texttt{Layout} objects. They all support the same set of transformation and operation APIs for maximum flexibility.} 
    \label{fig:api-specification}
  \end{figure}
\subsection{Layout Data Structures} 

A critical feature of \lp is the implementation of a series of data structures and operations that can be used to efficiently process and manipulate the layout elements. 
In document image analysis pipelines, various post-processing on the layout analysis model outputs is usually required to obtain the final outputs. 
Traditionally, this requires exporting DL model outputs and then loading the results into other pipelines. 
All model outputs from \lp will be stored in carefully engineered data types optimized for further processing, which makes it possible to build an end-to-end document digitization pipeline within \lp. 
There are three key components in the data structure, namely the \verb+Coordinate+ system, the \verb+TextBlock+, and the \verb+Layout+. 
They provide different levels of abstraction for the layout data, and a set of APIs are supported for transformations or operations on these classes.

Coordinates are the cornerstones for storing layout information. 
Currently, three types of \texttt{Coordinate} data structures are provided in \lp, shown in Figure~\ref{fig:api-specification}.
\verb+Interval+ and \verb+Rectangle+ are the most common data types and support specifying 1D or 2D regions within a document. They are parameterized with 2 and 4 parameters. 
A \verb+Quadrilateral+ class is also implemented to support a more generalized representation of rectangular regions when the document is skewed or distorted, where the 4 corner points can be specified and a total of 8 degrees of freedom are supported.
A wide collection of transformations like \verb+shift+, \verb+pad+, and \verb+scale+, and operations like \verb+intersect+, \verb+union+, and \verb+is_in+, are supported for these classes. 
Notably, it is common to separate a segment of the image and analyze it individually. 
\lp provides full support for this scenario via image cropping operations \verb+crop_image+ and coordinate transformations like \verb+relative_to+ and \verb+condition_on+ that transform coordinates to and from their relative representations. 
We refer readers to Table \ref{table:operations} for a more detailed description of these operations\footnote{This is also available in the \lp documentation pages.}.

Based on \texttt{Coordinate}s, we implement the \verb+TextBlock+ class that stores both the positional and extra features of individual layout elements. 
It also supports specifying the reading orders via setting the \texttt{parent} field to the index of the parent object.
A \verb+Layout+ class is built that takes in a list of \verb+TextBlock+s and supports processing the elements in batch. 
\verb+Layout+ can also be nested to support hierarchical layout structures.
They support the same operations and transformations as the \texttt{Coordinate} classes, minimizing both learning and deployment effort.

\begin{table}[t]
    \centering
    \caption{All operations supported by the layout elements. The same APIs are supported across different layout element classes including \texttt{Coordinate} types, \texttt{TextBlock} and \texttt{Layout}.}
    \label{table:operations}
    \resizebox{1.\linewidth}{!}{
    \begin{threeparttable}
        \setlength{\tabcolsep}{0.5em} 
        \renewcommand{\arraystretch}{1.2}
        \begin{tabular}{l|l}
            \toprule
            \textbf{Operation Name}             & \textbf{Description}                                                                                                                                \\

            \midrule
            \texttt{block.pad(top, bottom, right, left)} & Enlarge the current block according to the input \\
            \midrule
            \texttt{block.scale(fx, fy)}                 & \begin{tabular}[c]{@{}l@{}}Scale the current block given the ratio\\ in x and y direction\end{tabular}                                              \\
            \midrule
            \texttt{block.shift(dx, dy)}                 & \begin{tabular}[c]{@{}l@{}}Move the current block with the shift \\ distances in x and y direction\end{tabular}                                     \\
            \midrule
            \texttt{block1.is\_in(block2)}               & Whether block1 is inside of block2  \\
            \midrule
            \texttt{block1.intersect(block2)}            & \begin{tabular}[c]{@{}l@{}}Return the intersection region of block1 and block2.\\ Coordinate type to be determined based on the inputs.\end{tabular} \\
            \midrule
            \texttt{block1.union(block2)}                & \begin{tabular}[c]{@{}l@{}}Return the union region of block1 and block2. \\ Coordinate type to be determined based on the inputs.\end{tabular}    \\
            \midrule
            \texttt{block1.relative\_to(block2)}         & \begin{tabular}[c]{@{}l@{}}Convert the absolute coordinates of block1 to \\ relative coordinates to block2\end{tabular} \\
            \midrule
            \texttt{block1.condition\_on(block2)}        & \begin{tabular}[c]{@{}l@{}}Calculate the absolute coordinates of block1 given\\ the canvas block2's absolute coordinates\end{tabular} \\
            \midrule
            \texttt{block.crop\_image(image)}            & Obtain the image segments in the block region                        \\             
            \bottomrule
                                                                    
        \end{tabular}
    \end{threeparttable}
    }
\end{table}
\subsection{OCR} 

\lp provides a unified interface for existing OCR tools. 
Though there are many OCR tools available, they are usually configured differently with distinct APIs or protocols for using them. 
It can be inefficient to add new OCR tools into an existing pipeline, and difficult to make direct comparisons among the available tools to find the best option for a particular project. 
To this end, \lp builds a series of wrappers among existing OCR engines, and provides nearly the same syntax for using them. 
It supports a plug-and-play style of using OCR engines, making it effortless to switch, evaluate, and compare different OCR modules:

\begin{lstlisting}[language=python]
ocr_agent = lp.TesseractAgent() 
# Can be easily switched to other OCR software
tokens = ocr_agent.detect(image)
\end{lstlisting}

The OCR outputs will also be stored in the aforementioned layout data structures and can be seamlessly incorporated into the digitization pipeline. 
Currently \lp supports the Tesseract and Google Cloud Vision OCR engines. 

\lp also comes with a DL-based CNN-RNN OCR model~\cite{deng2017image} trained with the Connectionist Temporal Classification (CTC) loss~\cite{graves2006connectionist}. 
It can be used like the other OCR modules, and can be easily trained on customized datasets. 

\begin{figure}[t]
    \centering
    \includegraphics[width=0.80\linewidth]{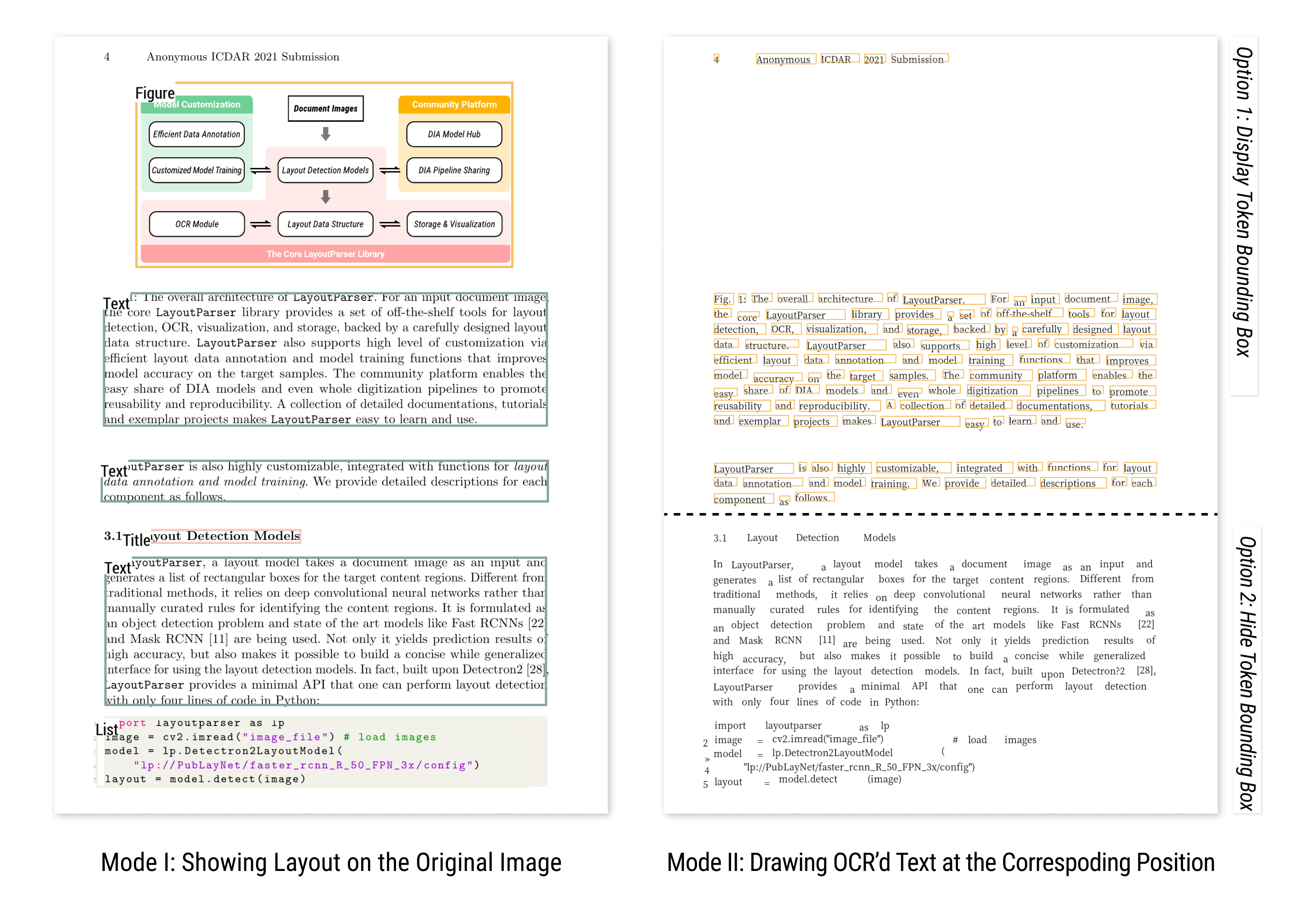}
    \caption{Layout detection and OCR results visualization generated by the \lp APIs. Mode I directly overlays the layout region bounding boxes and categories over the original image. Mode II recreates the original document via drawing the OCR'd texts at their corresponding positions on the image canvas. In this figure, tokens in textual regions are filtered using the API and then displayed.} 
    \label{fig:viz-example}
  \end{figure}
\subsection{Storage and visualization} 
The end goal of DIA is to transform the image-based document data into a structured database. 
\lp supports exporting layout data into different formats like \texttt{JSON}, \texttt{csv}, and will add the support for the METS/ALTO XML format~\footnote{https://altoxml.github.io} .
It can also load datasets from layout analysis-specific formats like COCO~\cite{zhong2019publaynet} and the Page Format~\cite{pletschacher2010page} for training layout models (Section~\ref{sec:training}). 

Visualization of the layout detection results is critical for both presentation and debugging. 
\lp is built with an integrated API for displaying the layout information along with the original document image. 
Shown in Figure~\ref{fig:viz-example}, it enables presenting layout data with rich meta information and features in different modes. 
More detailed information can be found in the online \lp documentation page.

\subsection{Customized Model Training}
\label{sec:training}

Besides the off-the-shelf library, \lp is also highly customizable with supports for highly unique and challenging document analysis tasks. 
Target document images can be vastly different from the existing datasets for training layout models, which leads to  low layout detection accuracy. 
Training data can also be highly sensitive and not sharable publicly. 
To overcome these challenges, \lp is built with rich features for efficient data annotation and customized model training. 

\lp incorporates a toolkit optimized for annotating document layouts using object-level active learning~\cite{shen2020olala}. 
With the help from a layout detection model trained along with labeling, only the most important layout objects within each image, rather than the whole image, are required for labeling.
The rest of the regions are automatically annotated with high confidence predictions from the layout detection model. 
This allows a layout dataset to be created more efficiently with only around 60\% of the labeling budget. 

After the training dataset is curated, \lp supports different modes for training the layout models. 
\emph{Fine-tuning} can be used for training models on a \emph{small} newly-labeled dataset by initializing the model with existing pre-trained weights. 
\emph{Training from scratch} can be helpful when the source dataset and target are significantly different and a large training set is available.
However, as suggested in Studer et al.'s work\cite{studer2019comprehensive}, loading pre-trained weights on large-scale datasets like ImageNet~\cite{imagenet_cvpr09}, even from totally different domains, can still boost model performance. 
Through the integrated API provided by \lp, users can easily compare model performances on the benchmark datasets. 

\section{\lp Community Platform}
\label{sec:community-platform}
Another focus of \lp is promoting the reusability of layout detection models and full digitization pipelines. 
Similar to many existing deep learning libraries, \lp comes with a community model hub for distributing layout models. 
End-users can upload their self-trained models to the model hub, and these models can be loaded into a similar interface as the currently available \lp pre-trained models. 
For example, the model trained on the News Navigator dataset~\cite{newspaper_navigator_dataset} has been incorporated in the model hub. 

Beyond DL models, \lp also promotes the sharing of entire document digitization pipelines.
For example, sometimes the pipeline requires the combination of multiple DL models to achieve better accuracy. 
Currently, pipelines are mainly described in academic papers and implementations are often not publicly available.
To this end, the \lp community platform also enables the sharing of layout pipelines to promote the discussion and reuse of techniques. 
For each shared pipeline, it has a dedicated project page, with links to the source code, documentation, and an outline of the approaches. 
A discussion panel is provided for exchanging ideas.
Combined with the core \lp library, users can easily build reusable components based on the shared pipelines and apply them to solve their unique problems.

\begin{figure}[t]
    \centering
    \includegraphics[width=\linewidth]{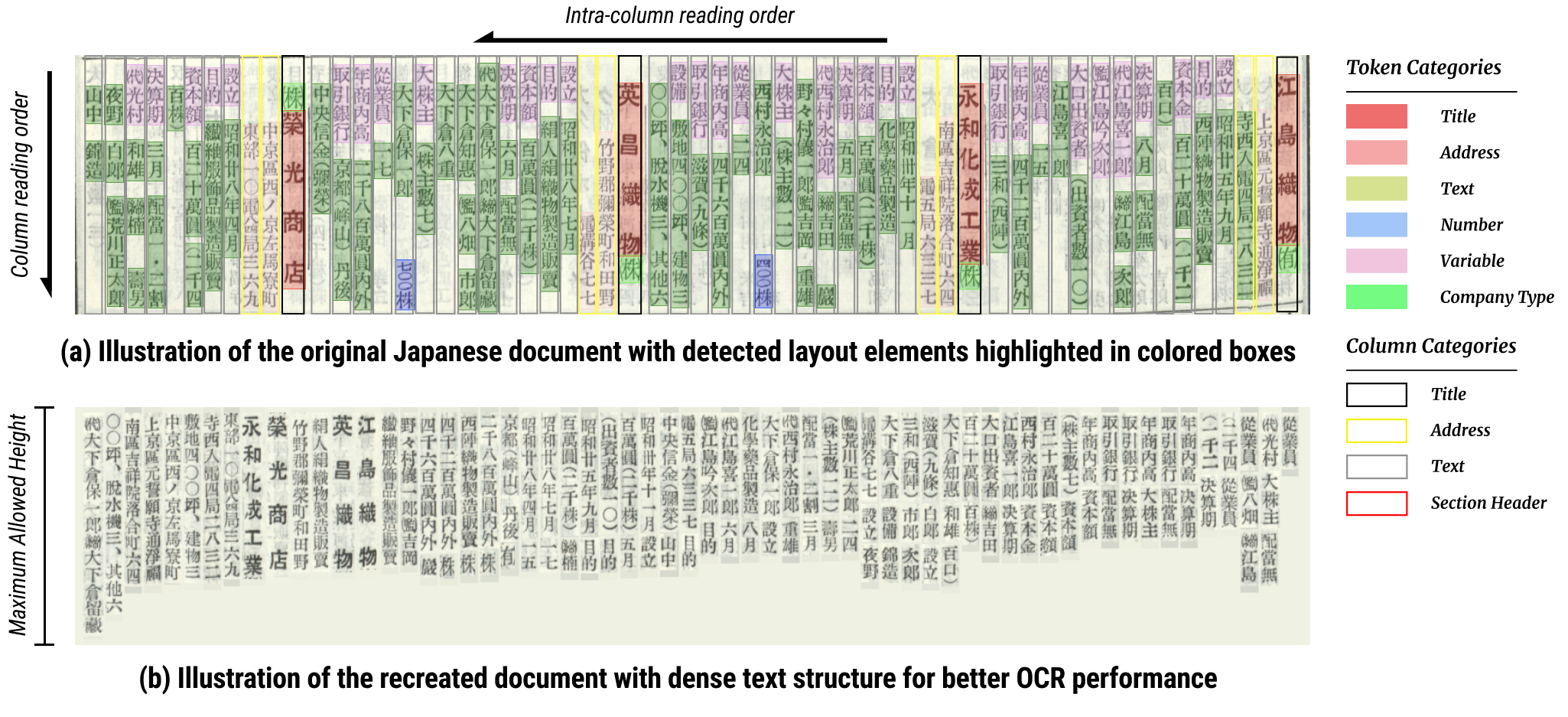}
    \caption{Illustration of (a) the original historical Japanese document with layout detection results and (b) a recreated version of the document image that achieves much better character recognition recall. The reorganization algorithm rearranges the tokens based on the their detected bounding boxes given a maximum allowed height.} 
    \label{fig:teikoku-example-combined}
\end{figure}

\section{Use Cases}
\label{sec:use-case}

The core objective of \lp is to make it easier to create both large-scale and light-weight document digitization pipelines. 
Large-scale document processing focuses on precision, efficiency, and robustness.
The target documents may have complicated structures, and may require training multiple layout detection models to achieve the optimal accuracy.
Light-weight pipelines are built for relatively simple documents, with an emphasis on development ease, speed and flexibility.
Ideally one only needs to use existing resources, and model training should be avoided. 
Through two exemplar projects, we show how practitioners in both academia and industry can easily build such pipelines using \lp and extract high-quality structured document data for their downstream tasks. 
The source code for these projects will be publicly available in the \lp community hub.

\subsection{A Comprehensive Historical Document Digitization Pipeline}

The digitization of historical documents can unlock valuable data that can shed light on many important social, economic, and historical questions. 
Yet due to scan noises, page wearing, and the prevalence of complicated layout structures, obtaining a structured representation of historical document scans is often extremely complicated. 
\begin{wrapfigure}{r}{0.5\textwidth}
    \begin{center}
    \includegraphics[width=\linewidth]{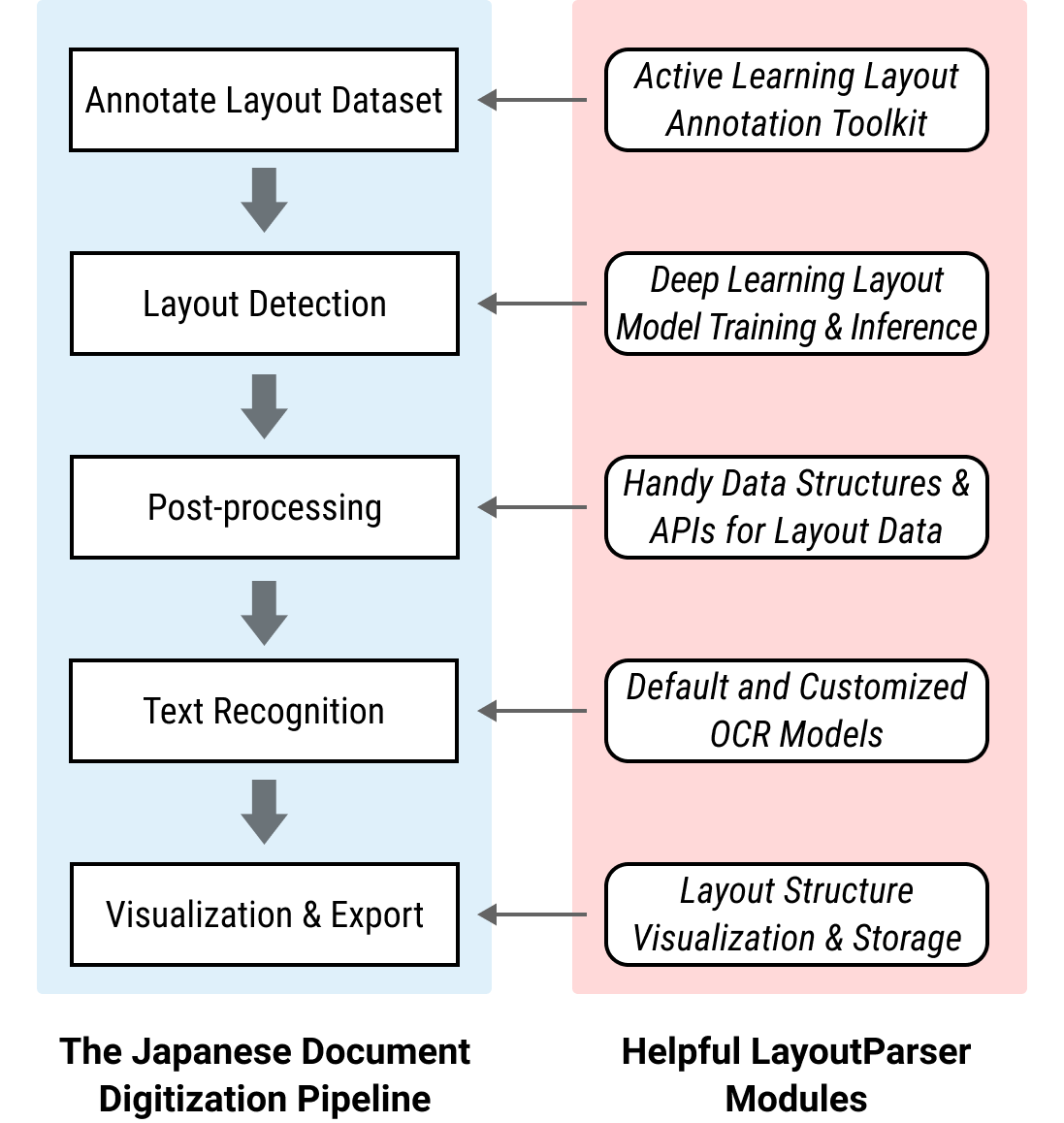}
    \end{center}
    \caption{Illustration of how \lp helps with the historical document digitization pipeline.} 
    \label{fig:teikoku-pipeline}
\end{wrapfigure}
In this example, \lp was used to develop a comprehensive pipeline, shown in Figure~\ref{fig:teikoku-pipeline}, to generate high-quality structured data from historical Japanese firm financial tables with complicated layouts.
The pipeline applies two layout models to identify different levels of document structures and two customized OCR engines for optimized character recognition accuracy. 

As shown in Figure~\ref{fig:teikoku-example-combined} (a), the document contains columns of text written vertically~\footnote{A document page consists of eight rows like this. For simplicity we skip the row segmentation discussion and refer readers to the source code when available.}, a common style in Japanese. 
Due to scanning noise and archaic printing technology, the columns can be skewed or have variable widths, and hence cannot be easily identified via rule-based methods.
Within each column, words are separated by white spaces of variable size, and the vertical positions of objects can be an indicator of their layout type.

To decipher the complicated layout structure, two object detection models have been trained to recognize individual columns and tokens, respectively. 
A small training set (400 images with approximately 100 annotations each) is curated via the active learning based annotation tool~\cite{shen2020olala} in \lp. 
The models learn to identify both the categories and regions for each token or column via their distinct visual features. 
The layout data structure enables easy grouping of the tokens within each column, and rearranging columns to achieve the correct reading orders based on the horizontal position. 
Errors are identified and rectified via checking the consistency of the model predictions. 
Therefore, though trained on a small dataset, the pipeline achieves a high level of layout detection accuracy: it achieves a 96.97 AP~\cite{lin2014microsoft} score across 5 categories for the column detection model, and a 89.23 AP across 4 categories for the token detection model. 

A combination of character recognition methods is developed to tackle the unique challenges in this document. 
In our experiments, we found that irregular spacing between the tokens led to a low character recognition recall rate, whereas existing OCR models tend to perform better on densely-arranged texts. 
To overcome this challenge, we create a document reorganization algorithm that rearranges the text based on the token bounding boxes detected in the layout analysis step.
Figure~\ref{fig:teikoku-example-combined} (b) illustrates the generated image of dense text, which is sent to the OCR APIs as a whole to reduce the transaction costs. 
The flexible coordinate system in \lp is used to transform the OCR results relative to their original positions on the page. 

Additionally, it is common for historical documents to use unique fonts with different glyphs, which significantly degrades the accuracy of OCR models trained on modern texts. 
In this document, a special flat font is used for printing numbers and could not be detected by off-the-shelf OCR engines. 
Using the highly flexible functionalities from \lp, a pipeline approach is constructed that achieves a high recognition accuracy with minimal effort.
As the characters have unique visual structures and are usually clustered together, we train the layout model to identify number regions with a dedicated category. 
Subsequently, \lp crops images within these regions, and identifies characters within them using a self-trained OCR model based on a CNN-RNN~\cite{deng2017image}. 
The model detects a total of 15 possible categories, and achieves a 0.98 Jaccard score\footnote{This measures the overlap between the detected and ground-truth characters, and the maximum is 1.} and a 0.17 average Levinstein distances\footnote{This measures the number of edits from the ground-truth text to the predicted text, and lower is better.} for token prediction on the test set.  

Overall, it is possible to create an intricate and highly accurate digitization pipeline for large-scale digitization using \lp. 
 The pipeline avoids specifying the complicated rules used in traditional methods, is straightforward to develop, and is robust to outliers. 
The DL models also generate fine-grained results that enable creative approaches like page reorganization for OCR. 

\begin{figure}[t]
    \centering
    \includegraphics[width=\linewidth]{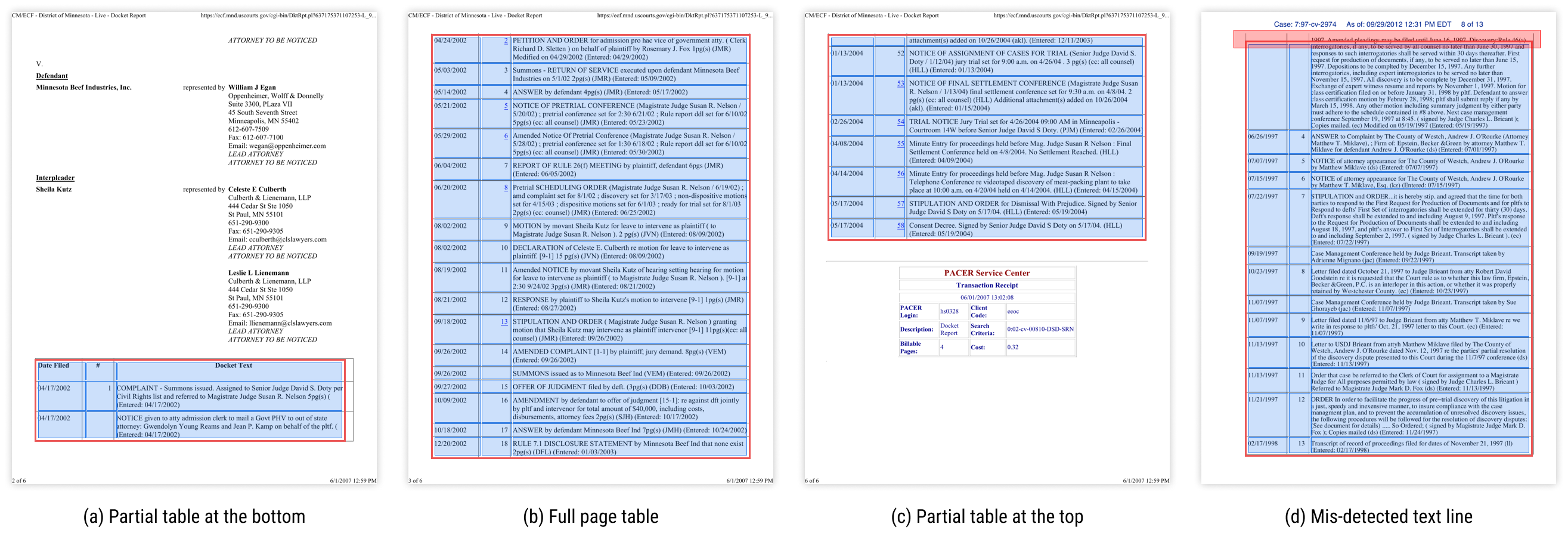}
    \caption{This lightweight table detector can identify tables (outlined in red) and cells (shaded in blue) in different locations on a page. In very few cases (d), it might generate minor error predictions, e.g, failing to capture the top text line of a table.} 
    \label{fig:table-detection}
  \end{figure}
\subsection{A light-weight Visual Table Extractor}

Detecting tables and parsing their structures (table extraction) are of central importance for many document digitization tasks. 
Many previous works~\cite{prasad2020cascadetabnet, schreiber2017deepdesrt, qasim2019rethinking} and tools~\footnote{https://github.com/atlanhq/camelot, https://github.com/tabulapdf/tabula} have been developed to identify and parse table structures. 
Yet they might require training complicated models from scratch, or are only applicable for born-digital PDF documents. 
In this section, we show how \lp can help build a light-weight accurate visual table extractor for legal docket tables using the existing resources with minimal effort. 

The extractor uses a pre-trained layout detection model for identifying the table regions and some simple rules for pairing the rows and the columns in the PDF image. 
Mask R-CNN~\cite{he2017mask} trained on the PubLayNet dataset~\cite{zhong2019publaynet} from the \lp Model Zoo can be used for detecting table regions. 
By filtering out model predictions of low confidence and removing overlapping predictions, \lp can identify the tabular regions on each page, which significantly simplifies the subsequent steps.  
By applying the line detection functions within the tabular segments, provided in the utility module from \lp, the pipeline can identify the three distinct columns in the tables. 
A row clustering method is then applied via analyzing the y coordinates of token bounding boxes in the left-most column, which are obtained from the OCR engines. 
A non-maximal suppression algorithm is used to remove duplicated rows with extremely small gaps.
Shown in Figure~\ref{fig:table-detection}, the built pipeline can detect tables at different positions on a page accurately. 
Continued tables from different pages are concatenated, and a structured table representation has been easily created. 

\section{Conclusion}
\label{sec:conclusion}

\lp provides a comprehensive toolkit for deep learning-based document image analysis.
The off-the-shelf library is easy to install, and can be used to build flexible and accurate pipelines for processing documents with complicated structures. 
It also supports high-level customization and enables easy labeling and training of DL models on unique document image datasets. 
The \lp community platform facilitates sharing DL models and DIA pipelines, inviting discussion and promoting code reproducibility and reusability. 
The \lp team is committed to keeping the library updated continuously and bringing the most recent advances in DL-based DIA, such as multi-modal document modeling~\cite{xu2019layoutlm, xu2020layoutlmv2,garncarek2020lambert} (an upcoming priority), to  a diverse audience of end-users.

\subsubsection*{Acknowledgements} We thank the anonymous reviewers for their comments and suggestions. This project is supported in part by NSF Grant OIA-2033558 and funding from the Harvard Data Science Initiative and Harvard Catalyst. Zejiang Shen thanks Doug Downey for suggestions. 

%
%
%
\bibliographystyle{splncs04}
\bibliography{main}
\end{document}